\documentclass[11pt,a4paper]{article}
\usepackage[hyperref]{acl2021}
\usepackage{times}
\usepackage{latexsym}
\usepackage{graphicx}
\usepackage{diagbox}
\usepackage{bm}
\usepackage{amsmath}
\usepackage{amssymb}
\usepackage{booktabs}
\usepackage{multirow}
\usepackage[footnotesize]{subfigure}

\newenvironment{shrinkeq}[1]
{ \bgroup
  \addtolength\abovedisplayshortskip{#1}
  \addtolength\abovedisplayskip{#1}
  \addtolength\belowdisplayshortskip{#1}
  \addtolength\belowdisplayskip{#1}}
{\egroup\ignorespacesafterend}
\usepackage{microtype}

\aclfinalcopy 

\title{Exploring Dynamic Selection of Branch Expansion Orders \\ for Code Generation}

\author{Hui Jiang\textsuperscript{1,2}\footnotemark[1]~~~Chulun Zhou\textsuperscript{1,2}\footnotemark[1]~~~Fandong Meng\textsuperscript{3}~~~Biao Zhang\textsuperscript{4}~~~Jie Zhou\textsuperscript{3}\\ \textbf{Degen Huang\textsuperscript{5}~~~Qingqiang Wu\textsuperscript{1,2}~~~Jinsong Su\textsuperscript{1,2}\footnotemark[2]}\\
  \textsuperscript{1}School of Informatics, Xiamen University \\
  \textsuperscript{2}Institute of Artificial Intelligence, Xiamen University\\
  \textsuperscript{3}Pattern Recognition Center, WeChat AI, Tencent Inc, China\\
  \textsuperscript{4}School of Informatics, University of Edinburgh~~~
  \textsuperscript{5}Dalian University of Technology\\
  \texttt{\small{\{hjiang,clzhou\}@stu.xmu.edu.cn}} ~~~
  \texttt{\small{\{fandongmeng,withtomzhou\}@tencent.com}}\\
  \texttt{\small{B.Zhang@ed.ac.uk}}~~~
  \texttt{\small{huangdg@dlut.edu.cn}}~~~
  \texttt{\small{\{wuqq,jssu\}@xmu.edu.cn}}
\thanks{Joint work with Pattern Recognition Center, WeChat AI, Tencent Inc, China.}
  }

\makeatletter
\def\thanks#1{\protected@xdef\@thanks{\@thanks
        \protect\footnotetext{#1}}}
\makeatother
\begin{document}
\maketitle
\renewcommand{\thefootnote}{\fnsymbol{footnote}}
\footnotetext[1]{Equal contribution}
\footnotetext[2]{Corresponding author}
\renewcommand{\thefootnote}{\arabic{footnote}}
\begin{abstract}
Due to the great potential in facilitating software development,
code generation has attracted increasing attention recently.
Generally, dominant models are Seq2Tree models, which convert the input natural language description into a sequence of tree-construction actions corresponding to the pre-order traversal of an Abstract Syntax Tree (AST).
However, such a traversal order may not be suitable for handling all multi-branch nodes.
In this paper,
we propose to equip the Seq2Tree model with a context-based \emph{Branch Selector},
which is able to dynamically determine optimal expansion orders of  branches for multi-branch nodes.
Particularly,
since the selection of expansion orders is a non-differentiable multi-step operation, we optimize the selector through reinforcement learning,
and formulate the
reward function as the difference of model
losses obtained through different expansion orders.
Experimental results and in-depth analysis on several commonly-used datasets demonstrate the effectiveness and generality of our approach.
We have released our code at \url{https://github.com/DeepLearnXMU/CG-RL}.
\end{abstract}

\section{Introduction}
Code generation aims at automatically generating a source code snippet given a natural language (NL)
description, which has attracted increasing attention recently due to its potential value in simplifying programming.
Instead of modeling the
abstract syntax tree (AST) of code snippets directly,
most of methods for code generation convert AST into a sequence of tree-construction actions.
This allows for using natural language
generation (NLG) models, such as the widely-used
encoder-decoder models, and obtains great success
\cite{Ling:2016ACL,Dong:2016ACL,Dong:2018ACL,Rabinovich:2017ACL,Yin:2017ACL,Yin:2018EMNLP,Yin:2019ACL,
Hayati:2018EMNLP,Sun:2019AAAI,Sun:2020AAAI,Wei:2019NIPS,Shin:2019NIPS,Xu:2020ACL,Binbin:2021AAAI}.
Specifically, an encoder is first used to learn word-level semantic representations of the input NL description.
Then, a decoder outputs a sequence of tree-construction actions, with which the corresponding AST is generated through pre-order traversal.
Finally, the generated AST is mapped into surface codes via certain deterministic functions.
%

\begin{figure}[t]
\centering
\includegraphics[width=7.5cm]{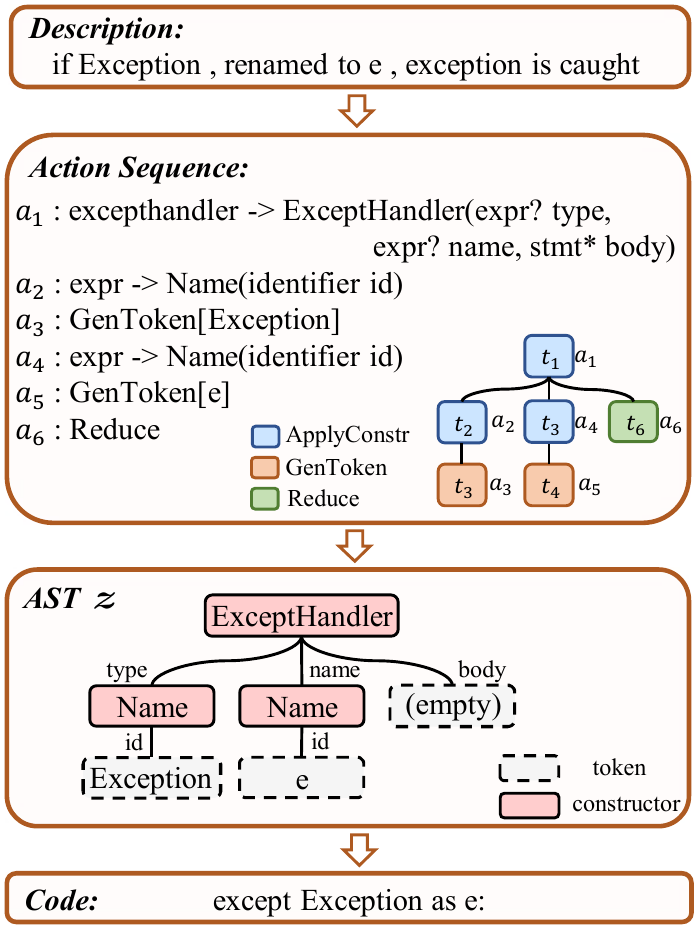}
\caption{An example of code generation using the conventional Seq2Tree model in pre-order traversal.}
\label{fig1}
\end{figure}

Generally,
during the generation of dominant Seq2Tree models based on pre-order traversal, branches of each multi-branch nodes are expanded in a left-to-right order.
Figure \ref{fig1} gives an example of the NL-to-Code conversion conducted by a Seq2Tree model.
At the timestep $t_1$,
the model generates a multi-branch node using the action $a_1$ with the grammar containing three fields: \emph{type}, \emph{name}, and \emph{body}.
Thus,
during the subsequent generation process,
the model expands the node of $t_1$ to sequentially generate several branches in a left-to-right order,
corresponding to the three fields of $a_1$.
The left-to-right order is a  conventional bias for most human-beings to handle multi-branch nodes,
which, however, may not be optimal for expanding branches.
Alternatively, if we first expand the field \emph{name} to generate a branch,
which can inform us the name `e',
it will be easier to expand the field \emph{type} with a `Exception' branch
due to the high co-occurrence of `e' and `Exception'.

To verify this conjecture,
we choose TRANX \cite{Yin:2018EMNLP} to construct a variant: TRANX-R2L, which conducts depth-first generation in a right-to-left manner, and then compare their performance on the DJANGO dataset.
We find that about 93.4\% of ASTs contain multi-branch nodes, and 17.38\% of AST nodes are multi-branch ones.
Table \ref{table1} reports the experimental results.
We can observe that
8.47\% and 7.66\% of multi-branch nodes can only be correctly handled by TRANX and TRANX-R2L, respectively.
Therefore, we conclude that different multi-branch nodes have different optimal branch expansion orders,
which can be dynamically selected based on context to improve the performance of conventional Seq2Tree models.

\begin{table}[t]
\centering
\begin{tabular}{lc}
  \toprule
     &  \textbf{Percentage} \\
  \midrule
  Only TRANX & 8.47  \\
  Only TRANX-R2L & 7.66 \\
  \bottomrule
\end{tabular}
\caption{
The percentages of multi-branch nodes, which can only be correctly handled by different models.
TRANX-R2L is a variant of TRANX \cite{Yin:2018EMNLP}, which handles multi-branch nodes in a right-to-left order.
}
\label{table1}
\end{table}

In this paper, we explore dynamic selection of branch expansion orders for code generation.
Specifically,
we propose to equip the conventional Seq2Tree model with a context-based \emph{Branch Selector},
which dynamically quantifies the priorities of expanding different branches for multi-branch nodes during AST generations.
However, such a non-differentiable multi-step operation poses a challenge to the model training.
To deal with this issue,
we apply reinforcement learning to train the extended Seq2Tree model.
Particularly,
we augment the conventional training objective with a reward function,
which is based on the model training loss between different expansion orders of branches.
In this way,
the model is trained to determine optimal expansion orders of branches for multi-branch nodes,
which will contribute to AST generations.

To summarize, the major contributions of our work are three-fold:
\begin{itemize}
\setlength{\itemsep}{0pt}
\setlength{\parsep}{0pt}
\setlength{\parskip}{0pt}
  \item Through in-depth analysis,
we point out that different orders of branch expansion are suitable for handling different multi-branch AST nodes,
and thus dynamic selection of branch expansion orders has the potential to improve conventional Seq2Tree models.
  \item We propose to incorporate a context-based \emph{Branch Selector} into the conventional Seq2Tree model
  and then employ reinforcement learning to train the extended model.
   To the best of our knowledge, our work is the first attempt to explore dynamic selection of branch expansion orders for code generation.
  \item Experimental results and in-depth analyses demonstrate the effectiveness and generality of our model on various datasets.
\end{itemize}

\section{Background}
As shown in Figure \ref{fig1}, the procedure of code generation can be decomposed into three stages.
Based on the learned semantic representations of the input NL utterance,
the dominant Seq2Tree model \cite{Yin:2018EMNLP} first outputs a sequence of abstract syntax description language (ASDL) grammar-based actions.
These actions can then be used to construct an AST following the pre-order traversal.
Finally,
the generated AST is mapped into surface code via a user-specified function \textup{AST\_to\_MR}($\ast$).

In the following subsections,
we first describe the basic ASDL grammars of Seq2Tree models.
Then, we introduce the details of TRANX \cite{Yin:2018EMNLP},
which is selected as our basic model due to its extensive applications and competitive performance \cite{Yin:2019ACL,Shin:2019NIPS,Xu:2020ACL}.
\footnote{Please note that our approach is also applicable to other Seq2Tree models.}

\subsection{ASDL Grammar}
%

Formally,
an ASDL grammar contains two components: \emph{type} and \emph{constructors}.
The value of \emph{type} can be composite or primitive.
As shown in the `$Action Sequence$' and `$AST \bm{z}$' parts of Figure \ref{fig1},
a constructor specifies a language component of a particular type using its fields, e.g., \emph{ExceptHandler (expr? type, expr? name, stmt$\ast$ body)}.
Each field specifies the type of its child node and contains a cardinality (single, optional ? and sequential $\ast$) indicating the number of child nodes it holds. For instance, \emph{expr? name} denotes the field \emph{name} has optional child node.
The field with composite type (e.g. \emph{expr}) can be instantiated by constructors of corresponding type,
while the field with primitive type (e.g. \emph{identifier}) directly stores token.

There are three kinds of ASDL grammar-based actions that can be used to generate the action sequence:
1) \textbf{\textsc{ApplyConstr}[$c$]}. Using this action, a constructor $c$ is applied to
the composite field of the parent node with the same type as $c$,
expanding the field to generate a branch ending with an AST node.
Here we denote the field of the parent node as \emph{frontier field}.
2) \textbf{\textsc{Reduce}}.
It indicates the completion of generating branches for a field with optional or multiple cardinalities.
3) \textbf{\textsc{GenToken}[$v$]}. It expands a
primitive frontier field to generate a token $v$.

Obviously, a constructor with multiple fields
can produce multiple AST branches\footnote{We also note that the field with sequential cardinality will be expanded to multiple branches. However, in this work, we do not consider this scenario, which is left as future work.}, of which generation order has important effect on the model performance, as previously mentioned.

\begin{figure*}[t]
\centering
\includegraphics[width=16cm]{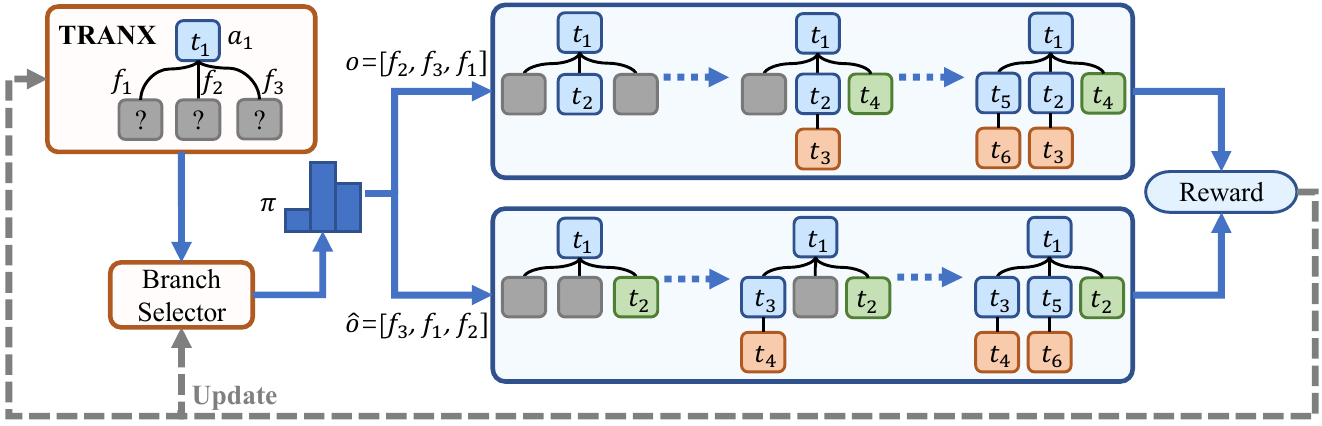}
\caption{The reinforced training of the extended TRANX model with branch selector.
We first fed the information of field and parent node into branch selector.
Then, from the policy probability distribution of branch selector,
we sample an order $o$ and infer an order $\hat{o}$.
Finally, we calculate the reward based on the model loss difference between $o$ and $\hat{o}$, and use the gradients to update parameters of the extended model.}
\label{fig2}
\end{figure*}

\subsection{Seq2Tree Model}
Similar to other NLG models, TRANX is trained to minimize the following objective function:
\begin{equation}
\mathcal{L}_{mle}(\bm{x}, \bm{a})=- \sum_{t=1}^{T} \log p(a_{t}|a_{<t}, \bm{x}),
\label{loss_mle}
\end{equation}
where $a_t$ is the $t$-th action,
and $p(a_t|a_{<t}, \bm{x})$ is modeled by an attentional encoder-decoder network \cite{Yin:2018EMNLP}.

For an NL description $\bm{x}$=${x_1,x_2,...,x_N}$,
we use a BiLSTM encoder to learn its word-level hidden states.
Likewise, the decoder is also an LSTM network.
Formally, at the timestep $t$, the temporary hidden state $\mathbf{h}_{t}$ is updated as
\begin{equation}
\mathbf{h}_{t}=f_{\mathrm{LSTM}}\left(\left[\mathbf{E}(a_{t-1}): \mathbf{s}_{t-1}: \mathbf{p}_{t}\right], \mathbf{h}_{t-1}\right),
\end{equation}
where $\mathbf{E}(a_{t-1})$ is the embedding of the previous action $a_{t-1}$,
$\mathbf{s}_{t-1}$ is the previous decoder hidden state,
and $\mathbf{p}_t$ is a concatenated vector involving
the embedding of the frontier field and
the decoder hidden state for the parent node.
Furthermore, the decoder hidden state $\mathbf{s}_{t}$ is defined as
\begin{equation}
\mathbf{s}_{t}=\tanh \left(\mathbf{W}\left[\mathbf{h}_{t}: \mathbf{c}_{t}\right]\right),
\end{equation}
where $\mathbf{c}_{t}$ is the context vector produced from the encoder hidden states
and $\mathbf{W}$ is a parameter matrix.

Here, we calculate the probability of action $a_t$ according to the type of its frontier field:
\begin{itemize}

\setlength{\itemsep}{0pt}
\setlength{\parsep}{0pt}
\setlength{\parskip}{0pt}
\item
\emph{Composite}. We adopt an \textsc{ApplyConstr} action to expand the field or a
\textsc{Reduce} action to complete the field.\footnote{\textsc{Reduce} action can be considered as a special \textsc{ApplyConstr} action}
The probability of using \textsc{ApplyConstr}[$c$] is defined as follows:
\begin{equation}
\begin{aligned}
&p\left(a_{t}\!=\!\operatorname{\textsc{ApplyConstr}}[c]|a_{<t}, \boldsymbol{x}\right)\\
=\ &\operatorname{softmax}\left(\mathbf{E}(c)^{\top} \mathbf{W} \mathbf{s}_{t}\right)
\end{aligned}
\end{equation}
where $\mathbf{E}(c)$ denotes the embedding of the constructor $c$.

\item
\emph{Primitive}. We apply a \textsc{GenToken} action to produce a token $v$,
which is either generated from the vocabulary or copied from the input NL description.
Formally, the probability of using \textsc{GenToken}[$v$] can be decomposed into two parts:
\begin{equation}\begin{aligned}
& p\left(a_{t}\!=\!\operatorname{\textsc{GenToken}}[v]|a_{<t}, \boldsymbol{x}\right) \\
=\ & p\left(\operatorname{gen} | a_{<t}, \boldsymbol{x}\right) p_{gen}\left(v | a_{<t}, \boldsymbol{x}\right)+ \\
&(1-p\left(\operatorname{gen} | a_{<t}, \boldsymbol{x}\right)) p_{copy}\left(v | a_{<t}, \boldsymbol{x}\right),
\end{aligned}
\label{predict_token}
\end{equation}
where $p\left(\operatorname{gen} | a_{<t}, \boldsymbol{x}\right)$ is modeled as
$\operatorname{sigmoid}\left(\mathbf{W} \mathbf{s}_{t} \right)$.

\end{itemize}
Please note that our proposed dynamic selection of branch expansion orders does not affect other aspects of the model.

\section{Dynamic Selection of Branch Expansion Orders}
In this section,
we extend the conventional Seq2Tree model with a context-based \emph{branch selector}, which dynamically determines optimal expansion orders of branches for multi-branch AST nodes.
In the following subsections,
we first illustrate the elaborately-designed branch selector module
and then introduce how to train the extended Seq2Tree model via reinforcement learning in detail.

\subsection{Branch Selector}
\label{sec:branch selector}
As described in Section 2.2,
the action prediction at each timestep is mainly affected by its previous action, frontier field and the action of its parent node.
Thus,
it is reasonable to construct the branch selector determining optimal expansion orders of branches according to these
three kinds of information.

Specifically,
given a multi-branch node $n_t$ at timestep $t$,
where the ASDL grammar of action $a_t$ contains $m$ fields $[f_1,f_2,...f_m]$,
we feed the branch selector with three vectors:
1) $\mathbf{E}(f_i)$: the embedding of field $f_i$,
2) $\mathbf{E}(a_t)$: the embedding of action $a_t$,
3) $\mathbf{s}_t$: the decoder hidden state,
and then calculate the priority score of expanding fields as follows:
\begin{equation}
Score(f_i)=\mathbf{W}_2(\tanh (\mathbf{W}_1[\mathbf{s}_{t}: \mathbf{E}(a_t): \mathbf{E}(f_i)])),
\end{equation}
where $\mathbf{W}_{1}$$\in$$\mathbb{R}^{d_{1} \times d_{2}}$ and
$\mathbf{W}_{2}$$\in$$\mathbb{R}^{d_{2} \times 1}$ are learnable parameters.\footnote{We omit the bias term for clarity.}

Afterwards, we normalize priority scores of expanding all fields into a probability distribution:
\begin{equation}
p_{n_t}= \operatorname{softmax}([Score(f_1):\cdots:Score(f_{m})]).
\end{equation}

Based on the above probability distribution, we can sample $m$ times to form a branch expansion order
$o = [f_{o_1},...,f_{o_m}]$, of which the policy probability is computed as
\begin{equation}
\pi(o)=\prod_{i=1}^{m}p_{n_t}(f_{o_i}|f_{o_{<i}}).
\end{equation}
It is notable that during the sampling of $f_{o_i}$, we mask previously sampled fields $f_{o_{<i}}$ to ensure that duplicate fields will not be sampled.

\subsection{Training with Reinforcement Learning}
During the generation of ASTs, with the above context-based branch selector, we deal with multi-branch nodes according to the dynamically determined order instead of the standard left-to-right order.
However,
the non-differentiability of multi-step expansion order selection and how to determine the optimal expansion order lead to challenges for the model training.
To deal with these issues, we introduce reinforcement
learning to train the extended Seq2Tree model in an end-to-end way.

Concretely, we first pre-train a conventional Seq2Tree model.
Then,
we employ self-critical training with a reward function that measures loss difference between
different branch expansion orders to train the extended Seq2Tree model.

\subsubsection{Pre-training}
It is known that a well-initialized network is very important for applying reinforcement learning \cite{Xiaomian:2020EMNLP}.
In this work,
we require the model to automatically quantify effects of different branch expansion orders on the quality of the generated action sequences.
Therefore,
we expect that the model has the basic ability to generate action sequences in random order at the beginning.
To do this,
instead of using the pre-order traversal based action sequences,
we use the randomly-organized action sequences to pre-train the Seq2Tree model.

Concretely,
for each multi-branch node in an AST,
we sample a branch expansion order from a uniform distribution,
and then reorganize the corresponding actions according to the sampled order.
We conduct the same operations to all multi-branch nodes of the AST, forming a new training instance.
Finally,
we use the regenerated training instances to pre-train our model.

In this way,
the pre-trained Seq2Tree model acquires the preliminary capability to make predictions in any order.

\subsubsection{Self-Critical Training}
With the above initialized parameters, we then perform self-critical training \cite{Steven:2017CVPR,Xiaomian:2020EMNLP} to update the Seq2Tree model with branch selector.

Specifically, we train the extended Seq2Tree model by combining the MLE objective
and RL objective together.
Formally,
given the training instance $(\bm{x}, \bm{a})$,
we first apply the sampling method described in section \ref{sec:branch selector} to all multi-branch nodes, reorganizing the initial action sequence $\bm{a}$ to form a new action sequence $\bm{a_o}$, and then define the model training objective as
\begin{shrinkeq}{-0.5ex}
\begin{equation}
\begin{aligned}
\mathcal{L} = \mathcal{L}_{mle}(\bm{a_{o}}|\bm{x};\theta) + \frac{\lambda}{|\textbf{N}_{mb}|} \sum_{n \in \textbf{N}_{mb}} \mathcal{L}_{rl}(o; \theta),
\end{aligned}
\end{equation}
\end{shrinkeq}
where $\mathcal{L}_{mle}(\ast)$ denotes the conventional training objective defined in Equation \ref{loss_mle}, $\mathcal{L}_{rl}(\ast)$ is the negative expected reward of
branch expansion order $o$ for the multi-branch node $n$,
$\lambda$ is a balancing hyper-parameter, $\textbf{N}_{mb}$ denotes the set of multi-branch nodes in the training instance,
and $\theta$ denotes the parameter set of our enhanced model.

More specifically,
$\mathcal{L}_{rl}(\ast)$ is defined as
\begin{shrinkeq}{-0.5ex}
\begin{equation}
\begin{aligned}
\mathcal{L}_{rl}(o; \theta)&=-\mathbb{E}_{o \sim \pi}[r(o)] \\
& \approx-r(o), o \sim \pi,
\end{aligned}
\end{equation}
\end{shrinkeq}
where we approximate the expected reward with the loss of an order $o$ sampled from the policy $\pi$.

Inspired by successful applications of self-critical training in previous studies \cite{Steven:2017CVPR,Xiaomian:2020EMNLP},
we propose the reward $r(\ast)$ to accurately measure the effect of any order on the model performance.
As shown in Figure \ref{fig2},
we calculate the reward using two expansion orders of branches: one is $o$ sampled from the policy $\pi$, and the other is $\hat{o}$ inferred from the policy $\pi$ with the maximal generation probability:
\begin{equation}
\begin{aligned}
r(o)=(\mathcal{L}_{mle}(\hat{o})-\mathcal{L}_{mle}(o))
*(\max(\eta-p(o),0)).
\end{aligned}
\end{equation}
Please note that we extend the standard reward function by setting a threshold $\eta$ to clip the reward, which can prevent the network from being over-confident in current expansion order of branches.

Finally, we apply the REINFORCE algorithm \cite{Ronald:1992} to compute the gradient:
\begin{equation}
\nabla_{\theta} \mathcal{L}_{rl} \approx-r\left(o\right) \nabla_{\theta} \log p_{\theta}\left(o\right).
\end{equation}

\begin{table*}[h]
\centering
\begin{tabular}{lcccc}
  \toprule
  \multirow{2}{*}{\textbf{Model}} & \textbf{DJANGO}  & \textbf{ATIS} & \textbf{GEO} & \textbf{CONALA} \\
   & Acc.  & Acc. & Acc. & BLEU / Acc. \\
  \midrule
   COARSE2FINE \cite{Dong:2018ACL}$^\dag$ 	 & --  & 87.7 & 88.2 & -- \\
   TRANX \cite{Yin:2019ACL}$^\dag$ 	 & 77.3 \scriptsize{$\pm$0.4} & 87.6 \scriptsize{$\pm$0.1} & 88.8 \scriptsize{$\pm$1.0} & 24.35 {\scriptsize{$\pm$0.4}} / 2.5 \scriptsize{$\pm$0.7} \\
   TREEGEN \cite{Sun:2020AAAI} & --  & 88.1 \scriptsize{$\pm$0.6} & -- & -- \\
   \midrule
  TRANX 	               & 77.2 \scriptsize{$\pm$0.6}  & 87.6 \scriptsize{$\pm$0.4}  & 88.8 \scriptsize{$\pm$1.0} & 24.38 {\scriptsize{$\pm$0.5}} / 2.2 \scriptsize{$\pm$0.5}\\
  TRANX (w/ pre-train)	   & 77.5 \scriptsize{$\pm$0.4}  &  87.8 \scriptsize{$\pm$0.7}  &  88.4\scriptsize{$\pm$1.1} & 24.57 {\scriptsize{$\pm$0.5}} / 1.4 \scriptsize{$\pm$0.3} \\
  TRANX-R2L             &   75.9 \scriptsize{$\pm$0.8}  &  87.5 \scriptsize{$\pm$0.9} & 86.4 \scriptsize{$\pm$1.0}  & 24.88 {\scriptsize{$\pm$0.5}} / 2.4 \scriptsize{$\pm$0.5}\\
  TRANX-RAND            & 74.6 \scriptsize{$\pm$1.1}  & 86.4 \scriptsize{$\pm$1.4} &  81.7 \scriptsize{$\pm$1.8} &  19.73 {\scriptsize{$\pm$1.1}} / 1.6 \scriptsize{$\pm$0.6} \\
   \midrule
  TRANX-RL (w/o pre-train) &  76.3 \scriptsize{$\pm$0.7} &   87.2  \scriptsize{$\pm$0.8}   &  87.1 \scriptsize{$\pm$1.6} & 23.38 {\scriptsize{$\pm$0.8}} / 2.1 \scriptsize{$\pm$0.2} \\
  TRANX-RL                 &   \textbf{77.9}  \scriptsize{$\pm$0.5}   & \textbf{89.1}  \scriptsize{$\pm$0.5}  &  \textbf{89.5} \scriptsize{$\pm$1.2} & \textbf{25.47} {\scriptsize{$\pm$0.7}} / \textbf{2.6} \scriptsize{$\pm$0.4}   \\
  \bottomrule
\end{tabular}
\caption{
The performance of our model in comparison with various baselines.
We report the mean performance and standard deviation over five random runs.
$^\dag$ indicates the scores are previously reported ones.
Note that we only report the result of TREEGEN on ATIS, since it is the only dataset with released code for preprocessing.
}
\label{main_result}
\end{table*}

\section{Experiments}
To investigate the effectiveness and generalizability of our model,
we carry out experiments on several commonly-used datasets.
\subsection{Datasets}
Following previous studies \cite{Yin:2018EMNLP,Yin:2019ACL,Xu:2020ACL}, we use the following four datasets:
\begin{itemize}
\setlength{\itemsep}{0pt}
\setlength{\parsep}{0pt}
\setlength{\parskip}{0pt}
  \item \textbf{DJANGO} \cite{Oda:2015ASE}. This dataset totally contains 18,805 lines of Python source code,
which are extracted from the Django Web framework, and each line is paired with an NL description.

  \item \textbf{ATIS}. This dataset is a set of 5,410 inquiries of flight information,
where the input of each example is an NL description and its corresponding output is a short piece of code
in lambda calculus.

  \item \textbf{GEO}. It is a collection of 880 U.S. geographical questions, with meaning representations
   defined in lambda logical forms like ATIS.

   \item \textbf{CONALA} \cite{Yin:2018MSR}. It totally consists of 2,879 examples of manually annotated NL questions and their Python solutions on STACK OVERFLOW. Compared
with DJANGO, the examples of CONALA
cover real-world NL queries issued by programmers with diverse intents, and are significantly more difficult due to its broad coverage
and high compositionality of target meaning
representations.
\end{itemize}

\subsection{Baselines}
To facilitate the descriptions of experimental results, we refer to the enhanced TRANX model as \textbf{TRANX-RL}.
In addition to TRANX, we compare our enhanced model with several competitive models:
\begin{itemize}
\setlength{\itemsep}{0pt}
\setlength{\parsep}{0pt}
\setlength{\parskip}{0pt}
  \item \textbf{TRANX (w/ pre-train)}. It is an enhanced version of TRANX with pre-training.
We compare with it because our model involves a pre-training stage.

  \item \textbf{COARSE2FINE} \cite{Dong:2018ACL}. This model adopts a two-stage decoding strategy to produce the action sequence. It first generates a rough sketch of its meaning, and then fills in missing detail.
  \item \textbf{TREEGEN} \cite{Sun:2020AAAI}. It introduces the attention mechanism of Transformer \cite{Vaswani:2017NIPS},
    and a novel AST reader to incorporate grammar and AST structures into the network.
  \item \textbf{TRANX-R2L}. It is a variant of the conventional TRANX model, which deals with multi-branch AST nodes in a right-to-left manner.
  \item \textbf{TRANX-RAND}. It is also a variant of the conventional TRANX model dealing with multi-branch AST nodes in a random order.

  \item \textbf{TRANX-RL (w/o pre-train)}.
In this variant of TRANX-RL, we train our model from scratch.
By doing so,
we can discuss the effect of pre-training on our model training.
\end{itemize}

To ensure fair comparisons, we use the same experimental setup as TRANX \cite{Yin:2018EMNLP}.
Concretely, the sizes of action embedding, field embedding and hidden states are set to 128, 128 and 256, respectively.
For decoding, the beam sizes for GEO, ATIS, DJANGO and CONALA are 5, 5, 15 and 15, respectively.
We pre-train models in 10 epochs for all datasets.
we determine the $\lambda$s as 1.0
according to the model performance on validation sets.
As in previous studies \cite{David:2017ICLR,Yin:2018EMNLP, Yin:2019ACL},
we use the exact matching accuracy (Acc) as the evaluation metric for all datasets.
For CONALA, we use the corpus-level BLEU \cite{Yin:2018MSR} as a complementary metric.

\begin{table*}[h]
\centering
\begin{tabular}{lcccc}
  \toprule
  \multirow{2}{*}{\textbf{Model}} & \textbf{DJANGO} & \textbf{ATIS} & \textbf{GEO}  & \textbf{CONALA}\\
   & Acc.  & Acc. & Acc. & Acc. \\
  \midrule
   TRANX 	& 77.26\scriptsize{$\pm$0.8} & 94.02\scriptsize{$\pm$0.8} & 89.75\scriptsize{$\pm$0.8} &  25.19\scriptsize{$\pm$0.6} \\
  TRANX-R2L &  76.88\scriptsize{$\pm$1.0} & 93.80\scriptsize{$\pm$0.3}  & 89.28\scriptsize{$\pm$1.1}   & 24.74\scriptsize{$\pm$0.7}   \\
  TRANX-RL  & \textbf{78.98}\scriptsize{$\pm$0.9}  & \textbf{94.87}\scriptsize{$\pm$0.5} & \textbf{90.64}\scriptsize{$\pm$0.9} &   \textbf{26.90}\scriptsize{$\pm$0.6}  \\
  \bottomrule
\end{tabular}
\caption{
Performance of our model in predicting actions for child nodes of multi-branch nodes.
}
\label{branch_result}
\end{table*}

\subsection{Main Results}
Table \ref{main_result} reports the main experimental results.
Overall, our enhanced model outperforms baselines across all datasets.
Moreover, we can draw the following conclusions:

First, our reimplemented TRANX model achieves comparable performance to previously reported results \cite{Yin:2019ACL} (TRANX).
Therefore, we confirm that our reimplemented TRANX model are convincing.

Second, compared with TRANX, TRANX-R2L and TRANX-RAND, our TRANX-RL exhibits better performance. This result demonstrates the advantage of dynamically determining branch expansion orders on dealing with multi-branch AST nodes.

Third, the TRANX model with pre-training does not gain a better performance.
In contrast,
removing the model pre-training leads to the performance degradation of our TRANX-RL model.
This result is consistent with the conclusion of previous studies \cite{Xin:2018ECCV,Xiaomian:2020EMNLP} that the pre-training is very important for the applying reinforcement learning.

\begin{table}[t]
\centering
\begin{tabular}{cccc}
  \toprule
   & \textbf{TRANX} & \textbf{TRANX-R2L}  & \textbf{TRANX-RL}  \\
  \midrule
  0 & 88.37 &\textbf{93.02} &90.11 \\
  1 & \textbf{100}& \textbf{100}& \textbf{100} \\
  2 & \textbf{100}& \textbf{100} & \textbf{100} \\
  3 & 78.94 & 81.57& \textbf{89.47}\\
  4 &\textbf{96.93} &\textbf{96.93} &\textbf{96.93} \\
  5 &\textbf{95.65}  & 95.23  &\textbf{95.65}  \\
  $\geq$6 & 78.75& 75.00& \textbf{80.63} \\
  \bottomrule
\end{tabular}

\caption{Accuracy on different data groups of ATIS according to the number of multi-branch nodes.}
\label{table_len_atis} 
\end{table}

\begin{table}[t]
\centering
\begin{tabular}{cccc}
  \toprule
   & \textbf{TRANX} & \textbf{TRANX-R2L}  & \textbf{TRANX-RL}  \\
  \midrule
  0 & \textbf{98.30} & 91.52 & 97.67 \\
  1 & \textbf{90.00}& \textbf{90.00}& \textbf{90.00} \\
  2 & 85.50 & 84.70 & \textbf{86.17} \\
  3 & 66.66 & 63.60 & \textbf{67.81}\\
  4 & 54.16 &48.33 & \textbf{57.50} \\
  5 & \textbf{28.88}  &26.66  &\textbf{28.88}  \\
  $\geq$6 & \textbf{12.35} & \textbf{12.35} & \textbf{12.35} \\
  \bottomrule
\end{tabular}

\caption{Accuracy on different data groups of DJANGO according to the number of multi-branch nodes.}
\label{table_len_dj} 
\end{table}

\subsection{Effects of the Number of Multi-branch Nodes}
As implemented in  related studies on other NLG tasks, such as machine translation \cite{Bahdanau:2015ICLR},
we individually split two relatively large datasets
(DJANGO and ATIS) into different groups according to the number of multi-branch AST nodes,
and report the performance of various models on these groups of datasets.

Tables \ref{table_len_atis} and \ref{table_len_dj} show the experimental results.
On most groups,
TRANX-RL achieves better or equal performance than other models.
Therefore,
we confirm that our model is general to datasets with different numbers of multi-branch nodes.

\subsection{ Accuracy of Action Predictions for the Child Nodes}
Given a multi-branch node, its child nodes have an important influence in the subtree. Therefore, we focus on the accuracy of action predictions for the child nodes.

For fair comparison,
we predict actions with previous ground-truth history actions as inputs.
Table \ref{branch_result} reports the experimental results.
We observe that TRANX-RL still achieves higher prediction accuracy than other baselines on most groups,
which proves the effectiveness of our model again.

\begin{figure}[h!]
\centering
\vspace{0.2cm}
\subfigure[The first example.]{
\captionsetup{font={tiny}}
\label{fig:subfig:1} 
\includegraphics[width=7.5cm]{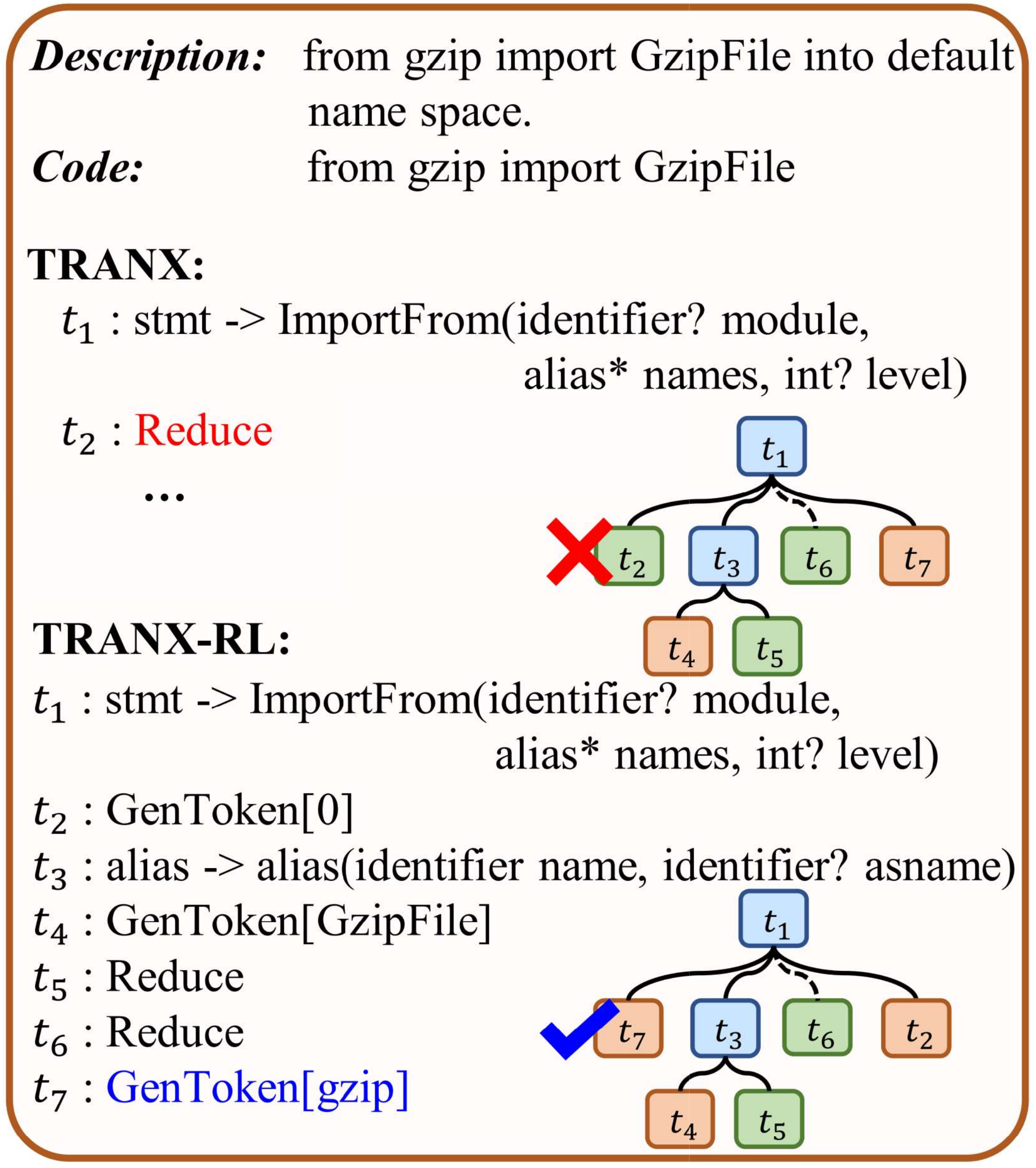}}
\subfigure[The second example.]{
\label{fig:subfig:2} 
\includegraphics[width=7.5cm]{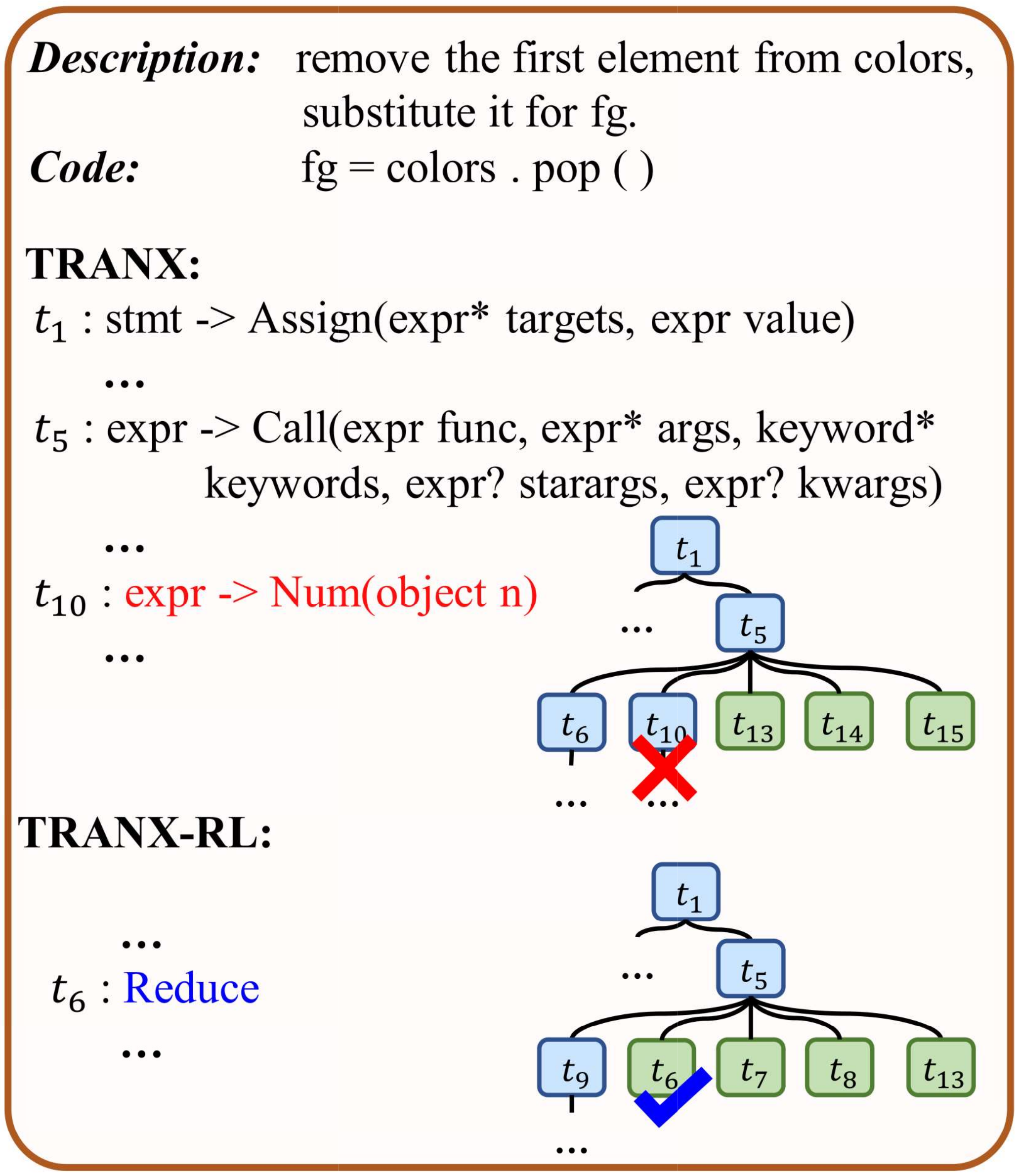}}
\caption{Two DJANGO examples produced by different models.}
\label{case} 
\end{figure}

\subsection{Case Study}
Figure \ref{case} shows two examples from DJANGO.
In the first example,
TRANX first generates the leftmost child node at the timestep $t_2$,
incorrectly predicting \textsc{GenToken}[`gzip'] as \textsc{Reduce} action.
By contrast, TRANX-RL puts this child node in the last position and successfully predict its action,
since our model benefits from the previously generated token `GzipFile' of the sibling node, which frequently occurs with `gzip'.

In the second example,
TRANX incorrectly predicts the second child node at the $t_{10}$-th timestep,
while TRANX-RL firstly predicts it at the timestep $t_6$.
We think this error results from the sequentially generated nodes and the errors in early timesteps would accumulatively harm the predictions of later sibling nodes.
By comparison, our model can flexibly generate subtrees with shorter lengths, alleviating error accumulation.

\section{Related Work}
With the prosperity of deep learning, researchers introduce neural networks into code generation.
In this aspect, \citet{Ling:2016ACL} first explore a Seq2Seq model for code generation.
Then, due to the advantage of tree structure,
many attempts resort to Seq2Tree models, which represent codes as trees of meaning representations \cite{Dong:2016ACL,David:2017ICLR,Rabinovich:2017ACL,Yin:2017ACL,Yin:2018EMNLP,Sun:2019AAAI,Sun:2020AAAI}.

Typically, \citet{Yin:2018EMNLP} propose TRANX, which introduces ASTs as intermediate representations of codes and has become the most influential Seq2Tree model.
Then, \citet{Sun:2019AAAI,Sun:2020AAAI} respectively explore CNN and Transformer architectures to model code generation.
Unlike these work,
\citet{Shin:2019NIPS} present a Seq2Tree model to
generate program fragments or tokens interchangeably at each generation step.
From another perspective, \citet{Xu:2020ACL} exploit external knowledge to enhance neural code generation model.
Generally,
all these Seq2Tree models generate ASTs in pre-order traversal, which, however, is not suitable to handle all
multi-branch AST nodes.
Different from the above studies that deal with multi-branch nodes in left-to-right order, our model determines the optimal expansion orders of branches for multi-branch nodes.

Some researchers have also noticed that the selection of decoding order has an important impact on the performance of neural code generation models.
For example,
\citet{David:2017ICLR} introduce a doubly RNN model that combines width and depth recurrences to traverse each node.
\citet{Dong:2018ACL} firstly generate a rough code sketch, and then fill in missing details by considering the input NL description and the sketch.
\citet{Jiatao:2019Trans} present an insertion-based Seq2Seq model that can flexibly
generate a sequence in an arbitrary order.
In general, these researches still deal with multi-branch AST nodes in a left-to-right manner.
Thus, these models are theoretically compatible with our proposed branch selector.

Finally, it should be noted that have been many NLP studies on exploring other decoding methods to improve other NLG tasks \cite{Xiangwen:2018AAAI,Jinsong:2019AI,Biao:2019TASLP,Sean:2019ICML,Mitchell:2019ICML,Jiatao:2019Trans,Jiatao:2019NIPS}.
However, to the best of our knowledge, our work is the first attempt to explore dynamic selection of branch expansion orders for tree-structured decoding.

\section{Conclusion and Future Work}
In this work,
we first point out that the generation of domainant Seq2Tree models based on pre-order traversal is not optimal for handling all multi-branch nodes. Then we propose an extended Seq2Tree model equipped with a context-based branch selector, which is capable of dynamically determining optimal branch expansion orders for multi-branch nodes.
Particularly, we adopt reinforcement learning to train the whole model with an elaborate reward that measures the model loss difference between different branch expansion orders.
Extensive experiment results and in-depth analyses demonstrate the effectiveness and generality of our proposed model
on several commonly-used datasets.

In the future, we will study
how to extend our branch selector to deal with indefinite branches caused by sequential field.

\section*{Acknowledgments}
The project was supported by National Key
Research and Development Program of China
(Grant No. 2020AAA0108004), National Natural Science
Foundation of China (Grant No. 61672440), Natural
Science Foundation of Fujian Province of China
(Grant No. 2020J06001), Youth Innovation Fund of Xiamen (Grant No. 3502Z20206059), and the Fundamental Research Funds for the Central Universities
(Grant No. ZK20720200077). We also thank the
reviewers for their insightful comments.

\bibliographystyle{acl_natbib}
\bibliography{refer,anthology}


\end{document}